# Inhomogeneous illumination image enhancement under extremely low visibility condition

**Libang Chen[1], Jinyan Lin[2], Qihang Bian[1], Yikun Liu[1,*] and Jianying Zhou[3]**

[1] Guangdong Provincial Key Laboratory of Quantum Metrology and Sensing & School of Physics and Astronomy, Sun Yat-Sen University (Zhuhai Campus), Zhuhai 519082, China

[2] School of Artificial Intelligence, Sun Yat-Sen University (Zhuhai Campus), Zhuhai 519082, China

[3] State Key Laboratory of Optoelectronic Materials and Technologies, School of Physics, Sun Yat-Sen University, Guangzhou 510275, China

[*] Correspondence: liuyk6@mail.sysu.edu.cn

**Abstract:** Imaging through dense fog presents unique challenges, with essential visual information crucial for applications like object detection and recognition obscured, thereby hindering conventional image processing methods. Despite improvements through neural network-based approaches, these techniques falter under extremely low visibility conditions exacerbated by inhomogeneous illumination, which degrades deep learning performance due to inconsistent signal intensities. We introduce in this paper a novel method that adaptively filters background illumination based on Structural Differential and Integral Filtering (SDIF) to enhance only vital signal information. The grayscale banding is eliminated by incorporating a visual optimization strategy based on image gradients. Maximum Histogram Equalization (MHE) is used to achieve high contrast while maintaining fidelity to the original content. We evaluated our algorithm using data collected from both a fog chamber and outdoor environments, and performed comparative analyses with existing methods. Our findings demonstrate that our proposed method significantly enhances signal clarity under extremely low visibility conditions and out-performs existing techniques, offering substantial improvements for deep fog imaging applications.

**Keywords:** Image enhancement, low visibility imaging, optical imaging, atmospheric scattering

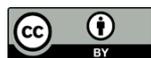

## 1. Introduction

An optical imaging system usually employs a point-to-point mapping between the object and the image to capture pristine information. However, under extremely low visibility condition, namely imaging through the atmosphere over long distance, the optical transmission process is disturbed, resulting in poor imaging quality. During transmission, atmospheric absorption results in the degradation of ballistic light. On the other hand, atmospheric scattering produces snake light and diffuse scattered light, which introduce noise and background interference to the original light field, respectively [1]. These complex processes are considered irreversible and entropy gaining [2, 3]. Thus, it has been a challenge to extract information over long distances through dense fog, or thick scattering media [4]. Yet for practical applications, extending the imaging distance through atmospheric scattering media to collect the desired light information is vital [5]. Although target recognition, object detection, or image enhancement techniques using deep neural network (DNN) have seen significant advances in recent years, the application of the DNN is often limited to certain atmosphere or weather conditions. In low visibility condition, the method require special image preprocessing before executing DNN [6-9].

To reconstruct image digitally under adverse weather conditions such as fog, haze, and sandstorms, numerous methods were developed [10–14]. The common idea across these methods is to suppress the scattered light and to recover information by utilizing the ballistic light, which contains the pristine information through the atmosphere. The comprehensive way is to amplify the captured image by a reasonable mapping, maintaining the fidelity while increasing its magnitudes, thus to enhance the perceptibility of a system. Histogram equalization (HE) is an example as an efficient way of mapping [15], utilizing the image's cumulative density function (CDF) to enhance visual contrast. The evolution of HE follows Contrast Clipped Adaptive histogram equalization (CLAHE) [16], which includes dividing image into fractions for localized HE and applying a bilinear interpolation between the boundaries of the blocks. The various algorithms based on CLAHE were subsequently proposed [17–19]. While these algorithms are quite successful in addressing the majority of scenarios involving atmospheric scattering, in situations with inhomogeneous illumination, the discontinuity of image blocks can nevertheless degrade image quality.

Generally, the inhomogeneity of illumination roots from the vignetting effects of the lenses, inconformity of the camera pixel response, and the inhomogeneity of illumination. In digital processing, methods such as top-hat transformation [20] and low-rank and sparse decomposition [21] can help achieve good effects in general. However, preserving details of the image for a single image input is still a challenge. The Single Scale Retinex (SSR) uses Gaussian kernel to filter the image, which is effective [22]. However, the method requires a manual selection of filter parameters. The Multi Scale Retinex (MSR) [23] solves the issue by applying the linear combination of various kernel size. Despite this improvement, the enhancement achieved is typically limited to linear stretching or exponential transformation. While MSR performs well under normal foggy conditions, its effectiveness diminishes significantly in extremely low visibility condition, where the reconstruction effects are often inadequate.

In a related study, Wang et al. [24] proposed a method that combines the Retinex theory, color space conversion, and adaptive logarithmic transformation to enhance low-light images. Although this method does not specifically address extremely low visibility conditions, it aims to improve overall image perception. Recently, deep learning-based image enhancement technique has become mainstream [25]. However, these methods can produce unexpected results, and are computationally expensive to train.

In this paper, we propose an effective image preprocessing technique, which outperforms existing image enhancement algorithms through thick scattering media in various ways. This enhancement technique is termed as Homogeneous Maximum Histogram Equalization (HMHE). To separate useful signals from background illumination, we introduce SDIF to evaluate the residual information in the image, thereby automatically selecting the appropriate filtering kernel to remove inhomogeneous illumination. Simultaneously, the grayscale banding effects caused by filtering are eliminated using a visual optimization algorithm based on the image gradient. Additionally, the filtered image is enhanced through MHE, which scales the desired components to their maximum contrast. Finally, the low and high frequencies of the image are linearly combined to complete the image information. Experiments have shown that the proposed method can fully utilize ballistic signals with high fidelity through scattering media, achieving fine visual perception of the desired targets. Due to the simplicity of the principle of our method, the output results are easily understood with confidence. Furthermore, the proposed method has a satisfying computational complexity, making it efficient and practical for real-world applications. The experimental results demonstrate that this method can serve as an effective image preprocessing technique for further neural network applications, such as image enhancement or target recognition.

To summarize, the key highlights of our paper are:

It produces high fidelity and contrast of the ballistic signals.

The straightforward principles underlying our method ensure that the results are easily interpretable, fostering confidence in its applications.

Our approach maintains a manageable computational complexity, making it suitable for integration into various image preprocessing pipelines for advanced neural network applications.

The paper is arranged as follows: In section 2, we briefly introduce the physical models in atmospheric imaging and the fundamental principles of HMHE. In section 3, the results in various scenes of HMHE are compared with existing similar methods, with qualitative results and discussion of the differences. The conclusions will be given in section 4.

## 2. Materials and Methods

### 2.1 Theory

#### 2.1.1 Low visibility imaging model

The optical process through scattering media is described in [26]. Due to the scattering effects, the reflected light from object is divided into the ballistic light, the snake light, which is slightly scattered, forming the noise component of the passing light, and the diffuse scattered light, which suffers multiple scattering, and becomes a global interference. We assume the atmospheric extinction coefficient $\beta_{ext}$ a constant over space and time. Therefore, we can write:

$$I_{view} = I_{bal} + I_{sna} + I_{sca} = I_{obj}T + I_{sna} + I_{illu}(1 - T) \tag{1}$$

where $I_{view}$ is the viewed light, $I_{bal}$ is the ballistic light, $I_{sna}$ is the snake light, $I_{sca}$ is the diffuse scattered light, and $I_{obj}$ is the directly reflected light from the object. The ballistic light follows geometric optics, and suffers degradation by the Lambert-Beer law as $T = \exp(-\beta_{ext}L)$. It can be seen from Eq. (1) that the ballistic light has a weak amplitude, while the scattered light has a strong amplitude, resulting in the signal with a property of low contrast. Meanwhile, the snake light renders the signal a low signal-to-noise ratio.

#### 2.1.2 Illumination correction

We denote the imaged intensity with inhomogeneous illumination as the simple sum of the below quantities [27]:

$$U(x,y) = O(x,y) + B(x,y) + N(x,y) \tag{2}$$

where $O(x,y)$, $B(x,y)$, and $N(x,y)$ are the image of the object, the illumination, and the additive noise. $x$, $y$ are the transverse and vertical positions. We assume the illumination is of low frequency term [21], and the desired target at a distance is of high frequency. By applying an appropriate low pass filter (LPF), we achieve the desired low frequency components of the total image, or the illumination $U_{illu}$:

$$U_{illu}(x,y) = O_{low}(x,y) + B(x,y) \tag{3}$$

Thus, the image with only the useful information can be expressed as：

$$U_{homo}(x,y) = U(x,y) - U_{illu}(x,y) = O_{high}(x,y) + N(x,y) \tag{4}$$

The transformation from light to digits is denoted as $f$. In expression of the light field:

$$U_{homo}(x,y) = f(I_{obj}^{high}T + I_{sna}) \tag{5}$$

By enhancing the homogeneous components of the original image, the interference of the illumination is eliminated.

#### 2.1.3 MHE

HE is a simple image enhancement algorithm to enhance the original contrast by means of projecting intensity by the CDF. MHE is a variation of HE, which is proposed here to fully expand the histogram distribution of an image.

Assuming the allowed discrete value $\mathbb{Z} \in [0, L-1]$, where $L$ is the maximum grayscale of the camera. The histogram of an image is $H(i) = n(i)/MN$, where $i$ represents one of the values in $\mathbb{Z}$, and $H(i)$ represents the number of pixels which has the intensity $i$. $M$, $N$ are the number of pixels in vertical and horizonal direction respectively. The CDF of is denoted as $D(i) = \sum_{k=0}^{i} H(k)/MN$.

We denote the enhanced intensity as $e$. We propose MHE by the above equation:

$$e = (L-1)\frac{\alpha[D(i) - D(i_{min})]}{D(i_{max}) - D(i_{min})} \tag{6}$$

where $\alpha$ is a scaling factor applied to control the balance between the amplitude of the signal and illumination. The enhancement technic effectively amplifies the original signal with high fidelity by keeping the original signal value relationship following the CDF.

### 2.1.4 Contrast of the enhanced image

The contrast of an object is defined as:

$$C = \frac{i_{bright} - i_{dark}}{i_{bright} + i_{dark}} \tag{7}$$

We denote the histogram of the inhomogeneous illuminated image as $H'(i)$, the object and background intensity $i_{bright}$ and $i_{dark}$ respectively. The corrected histogram is $H(i)$. For $\alpha = 1$, we can write the contrast of the HE and HMHE enhanced image as:

$$C_{HE} = \frac{\Sigma_{k=0}^{i_{bright}} H'(k) - \Sigma_{k=0}^{i_{dark}} H'(k)}{\Sigma_{k=0}^{i_{bright}} H'(k) + \Sigma_{k=0}^{i_{dark}} H'(k)} \tag{8}$$

$$C_{HMHE} = \frac{\Sigma_{k=0}^{i_{bright}} H(k) - \Sigma_{k=0}^{i_{dark}} H(k)}{\Sigma_{k=0}^{i_{bright}} H(k) + \Sigma_{k=0}^{i_{dark}} H(k) - 2H(i_{min})} \tag{9}$$

It can be seen for Eq (8), (9) that the contrast of the HMHE enhanced image incorporates a negative term in the dominator, thus compared to the HE enhanced image, assuming identical histograms for both, the amplification of contrast is significant. Fig. (1) (a), (b), (c), and (d) demonstrates the original, enhanced, filtered and HMHE enhanced image histogram of the contaminated image respectively. Owing to the illumination, the histogram is inevitably widened, leading to a weakened enhancement space for signal, and an unwanted enhancement of the illumination, leading to poor image quality.

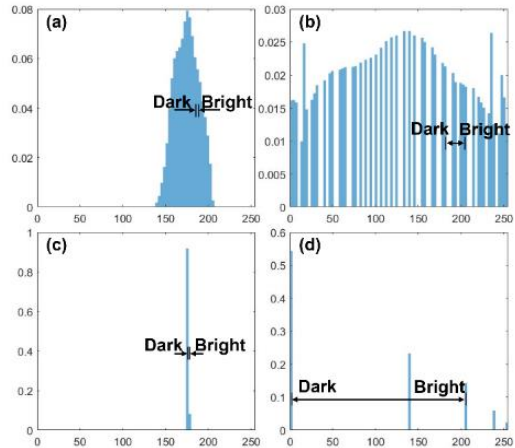

**Fig. 1**. An illustration of the image histogram. (a) histogram in low visibility condition with inhomogeneous illumination. (b) HE enhanced histogram for (a). (c) filtered histogram in low of (a). (d) MHE enhanced histogram for (a).

### 2.2 HMHE

#### 2.2.1 Overview

The HMHE consists of three parts. The background light extraction, the visual enhancement, and the image enhancement. A flow chart of the algorithm is given in Fig. 2. In the background light extraction module, an estimator based on Structure Similarity Index Measurement (SSIM) [28] is proposed to assure the filtering of only the background light, remaining the desired signals. In the visual perception optimization module, it addresses the grayscale banding introduced due to applying a LPF to the image. In the image enhancement module, the low and high frequency components of the image are enhanced separately, and recombined as the output.

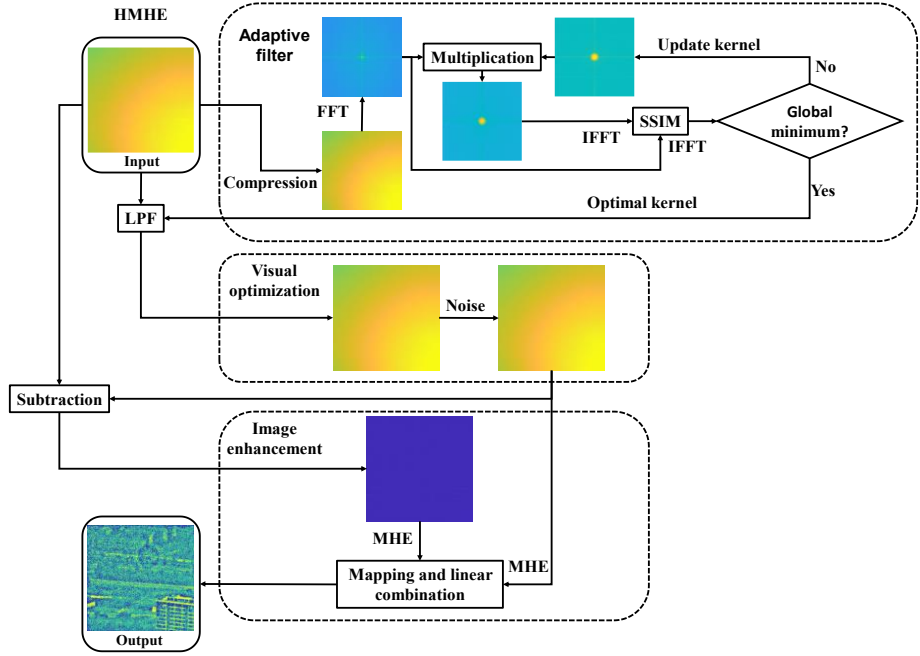

**Fig. 2**. The overview flow chart of HMHE.

The computational complexity of HMHE can be expressed as:

$$O\big(MNlog(MN)\big) \tag{10}$$

### 2.2.2 SDIF

Here, we propose an effective method to filter the optimal target for a desired image. For images corrupted by inhomogeneous illumination, estimating the background light is crucial. Based on the assumption that illumination is a low-frequency signal, LPF is applied to estimate and remove the it. However, for complex scenes, the selection of filter parameters is critical. An adaptive algorithm based on image features to automatically estimate the appropriate kernel, and effectively removing inhomogeneous illumination while preserving image details is proposed.

A windowed Gaussian function is chosen for its naturalness in signal processing [29] as the convolution kernel of the LPF. The standard variance $\sigma$ of the Gaussian function and its kernel size satisfies $k = 2\lceil 2\sigma \rceil + 1$. Here, $\lceil \cdot \rceil$ means rounding up to the nearest integer. To prevent negative values, we subtract the original image's minimal value from the filtered image.

$$I_{filtered}(x,y) = I_{LPF}(x,y) - \min[I(x,y)] \tag{11}$$

The expression for SSIM of the original image and the filtered image is given by:

$$SSIM(X,Y) = \frac{2\mu_X\mu_Y(2\sigma_{XY} + C_2)}{(\mu_X^2 + \mu_Y^2 + C_1)(\sigma_X^2 + \sigma_Y^2 + C_2)} \tag{12}$$

where $X$, $Y$ are the original and filtered image, $\mu_{X,Y}$, $\sigma_{X,Y}$ are the average and variances of $X,Y$, $\sigma_{xy}$ is the covariance, $C_1 = (K_1L)^2$, $C_2 = (K_2L)^2$ are the regulation constants, $K_1$, $K_2$ are two small constants. SSIM has sufficient resolution to judge the structural similarity between two images. In our study, the determination of the cutoff kernel $k_{cutoff}$ is given by the following expression:

$$k_{cutoff} = P * \min\big[\big|\sum SSIM(I, I_{filtered})\big|\big] + I * \min\big[\big|SSIM(I, I_{filtered})\big|\big] + D * \min\big[\big|\nabla_k SSIM(I, I_{filtered})\big|\big] \tag{13}$$

where $P = 1$, $I = 0.3$, $D = 0.7$ are respectively three empirical coefficients, $|\cdot|$, $\nabla_k$ denote taking the absolute and derivative of the variable $k$. The adjustment of $k_{cutoff}$ has instrumental impact on image quality. Initially, as the kernel begins to exclude the detailed and effective information, SSIM experiences a notable decrease. This stage highlights the importance of the derivatives: the first derivative captures how swiftly structural similarity diminishes. Once the filter removes all effective information, further enlarging the kernel impacts SSIM minimally, as the remaining low-frequency components do not significantly affect the image's structural similarity. The optimal balance between eliminating irrelevant noise and preserving vital structural features is achieved by applying the weighted combination of the integral, original, and derivative of SSIM, ensuring the selected cutoff kernel maintains image quality by keeping significant structural information intact.

Fig. 3(a), (b), and (c) demonstrate a high-information scenario, showing the original image, the filtered result using our algorithm, and the SSIM relationship with the kernel, respectively. Similarly, Fig. 3(d), (e), and (f) depict a low-information scenario. In the first scenario, the useful information consists mainly of high-frequency components, resulting in a kernel size of 48. In contrast, the second scenario, where the target consists of two floaters in the ocean and contains lower-frequency components, requires a larger kernel size of 128 to fully filter the information.

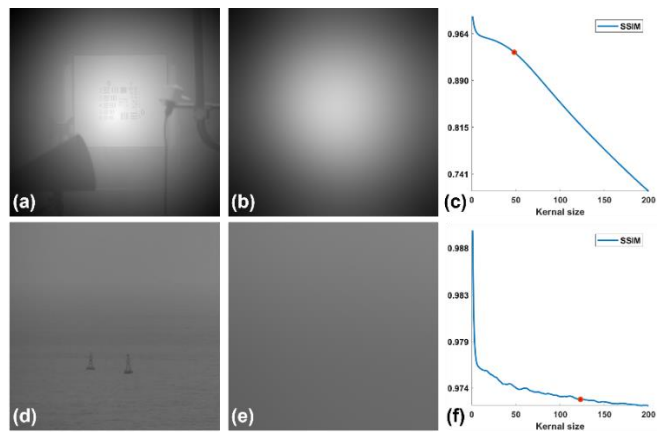

**Fig. 3**. A demonstration of the filtering selection algorithm. (a), (d) original image, (b), (e) filtered image, (c), (f) the SSIM relation with the kernel size.

### 2.2.3 Visual optimization

Applying LPF to an image contributes to the grayscale banding effect. If this phenomenon is not addressed, the final enhanced image displays poor visual perception. The reason for this effect is that the low frequency components of the image decrease in a stair-stepping manner in the spatial domain, leading to contour-like images. Moreover, when performing global mapping, these bandings will be amplified. Luckily, potential attempts such as noise addition proves to suppress such phenomenon [30]. In order to eliminate the banding effect without affecting the image quality, we eliminate the artefacts by adding a minimum interference to the bands:

$$I_{filtered}^{visual}(x, y) = I_{filtered}(x + m, y + n) + N_{add}(x + m, y + n) \tag{14}$$

where $I_{filtered}^{visual}$ is the optimized image. $N_{add}$ is the added gaussian noise whose standard variance satisfies $\sigma_n = \max[gradient(I_{filtered})]/3$. The comparison between the image before and after the visual optimization algorithm are shown in Fig. 4, where Fig. 4(a)-(d) demonstrates the original filtered images whereas Fig. 4(e)-(h) are the filtered images after the visual optimization algorithm. Specifically, Fig. 4(a), (d) are the processing results in the fog chamber, while Fig. 4(b), (c) are outdoor natural low-visibility scenes. It can be seen from the results that the banding effect is visually resolved after applying the visual optimization algorithm.

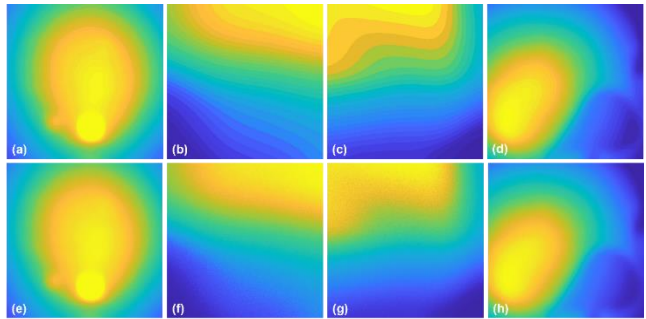

**Fig. 4.** Comparison of filtered images before and after the visual optimization algorithm in different scenarios. The laboratory environment images are placed on the left and right sides (a, d) for the original images and (e, h) for the processed images. The outdoor natural low-visibility scenes are placed in the middle (b, c) for the original images and (f, g) for the processed images.

### 2.2.4 Image enhancement

The image enhancement technic in this work is the weighted combination of both filtered and unfiltered images through MHE. When images are filtered by a large kernel, the low frequency components are completely removed. However, by linear combining both the low and high frequency components of the image, all the components in the frequency domain would be preserved, and the artifacts introduced by filtering can be deduced. Here, we apply the image enhancement by utilizing both MHE and a linear combination of high and low frequency components. MHE achieves the maximum available contrast of the original image following the rules of HE, leading to high contrast. And the combination allows image to preserve more details and image fidelity. The expression for the image enhancement technic is expressed as follows:

$$e = (L-1)\left\{\frac{\alpha[D_{homo}(i) - D_{homo}(i_{min})]}{[D_{homo}(i_{max}) - D_{homo}(i_{min})]} + \frac{(1-\alpha)[D_{illu}(i) - D_{illu}(i_{min})]}{[D_{illu}(i_{max}) - D_{illu}(i_{min})]}\right\}, \tag{15}$$

where $\alpha$ is the coefficient that adjust the proportion between the high and low frequency components of the image.

## 3. Results and Discussions

### 3.1. Experimental setup

To thoroughly verify the effectiveness of the algorithm proposed in this paper, experiments were conducted in both indoor and outdoor settings. Thus, our data encompasses scenarios involving indoor lighting and artificial fog, as well as natural sunlight and natural fog. To rigorously assess the robustness of the algorithm, experiments were carried out using cameras and lenses of various models. The cameras used in the experiments include the DAHENG MER-232-48GM-P-NIR, PCO EDGE 4.2, and Hamamatsu C11440-22CU, with resolutions of 1920x1200 pixels, 8-bit depth; 2048x2048 pixels, 16-bit depth; and 2048x2048 pixels, 16-bit depth, respectively. The lenses employed were Computar M7528-MP and Zeiss Milvus 2/100M.

The indoor experiments were conducted at the Visibility Calibration Laboratory of the China Meteorological Administration in Shanghai, China. We used targets of varying spatial frequencies, shapes, and distances to mimic real-world natural scenes. Due to the controllable fog density in the indoor fog chamber, we were able to obtain clear images without fog, which offers a significant advantage for evaluating the algorithm. The data from outdoor experiments were used to further validate the reliability of the algorithm.

### 3.2 Evaluation metrices

For quantitative evaluation, we selected the following Image Quality Assessment (IQA) metrics: Information Entropy (IE), SSIM, Feature Similarity Index (FSIM) [31], and correlation. These metrics allow us to comprehensively assess various aspects of image quality, including detail richness, similarity, detail similarity and correlation to reference images.

1) IE

IE evaluates the average entropy of an image. A high entropy indicates distinct structure and detail richness [32]. The definition of IE for any given image is:

$$IE = -\Sigma p_i \log_2 p_i, \tag{16}$$

where $p_i$ is the probability of a particular gray level in the image.

2) SSIM

SSIM is utilized to assess the similarity between the processed images and their references, which may be the ground truth or the original images, depending on the available dataset. This metric, previously detailed, focuses on measuring the visual impact of three characteristics of an image: luminance, contrast, and structure, thereby providing a comprehensive evaluation of image quality across different types of comparisons.

3) FSIM

FSIM is defined as follows:

$$FSIM = \frac{\Sigma S_L * PC_m}{\Sigma PC_m}, \tag{17}$$

Where $S_L$ stands for similarity, $PC_m$ is the phase congruency. This value gives the low-level features of an image.

4) Correlation

The correlation of two images can be calculated by

$$Corr = \frac{\Sigma(I - \mu_I)(K - \mu_K)}{\sqrt{\Sigma(I - \mu_I)\Sigma(K - \mu_k)}}, \tag{18}$$

This value reflects the correlation between two images. The higher the value, the more related the two images are.

### 3.3 Comparisons of image enhancement methods

In this paper, we compare the proposed algorithm with identical algorithms specially designed for image enhancement through fog, including CLAHE, SSR, and Adaptive Multi Scale Retinex (AMSR) [33]. For CLAHE, the clip limit for normalized images is set to 0.02, with image segmentation into 8×8 blocks. The SSR uses a Gaussian function of standard deviation set to 50 pixels. The experimental data from the fog chamber and the comparison results are illustrated in Fig. 5, where (a)-(f) depict multiple targets inside the fog chamber. The upmost row shows data captured in clear conditions. The second to sixth rows respectively showcase the original low visibility image, the outcomes of CLAHE, SSR, AMSR and HMHE.

The data analysis indicates that in indoor environments characterized by severe uneven lighting, CLAHE struggles with significant light intensity variations between blocks, resulting in conspicuous block artifacts due to its inability to harmonize the light intensity across different tiles effectively. SSR, which incorporates homomorphic filtering, successfully recovers image information; however, it yields a subdued contrast that obscures the visibility of the signal. AMSR shows performance comparable to our algorithm. Yet, the both CLAHE, SSR, and AMSR over amplifies the noise in the high frequency domain.

Quantitative evaluations presented in Table 1, corresponding to Fig. 5, demonstrate that while the algorithm proposed in this work does not always outperform competing methods, it achieves competitive results across various metrics. However, despite these numerical comparisons, the visual perception of images processed by our algorithm is subjectively superior, suggesting enhanced viewer satisfaction. This subjective improvement in visual quality is crucial, as it potentially facilitates easier feature extraction for neural networks, thus improving the efficiency and accuracy of subsequent automated analysis tasks.

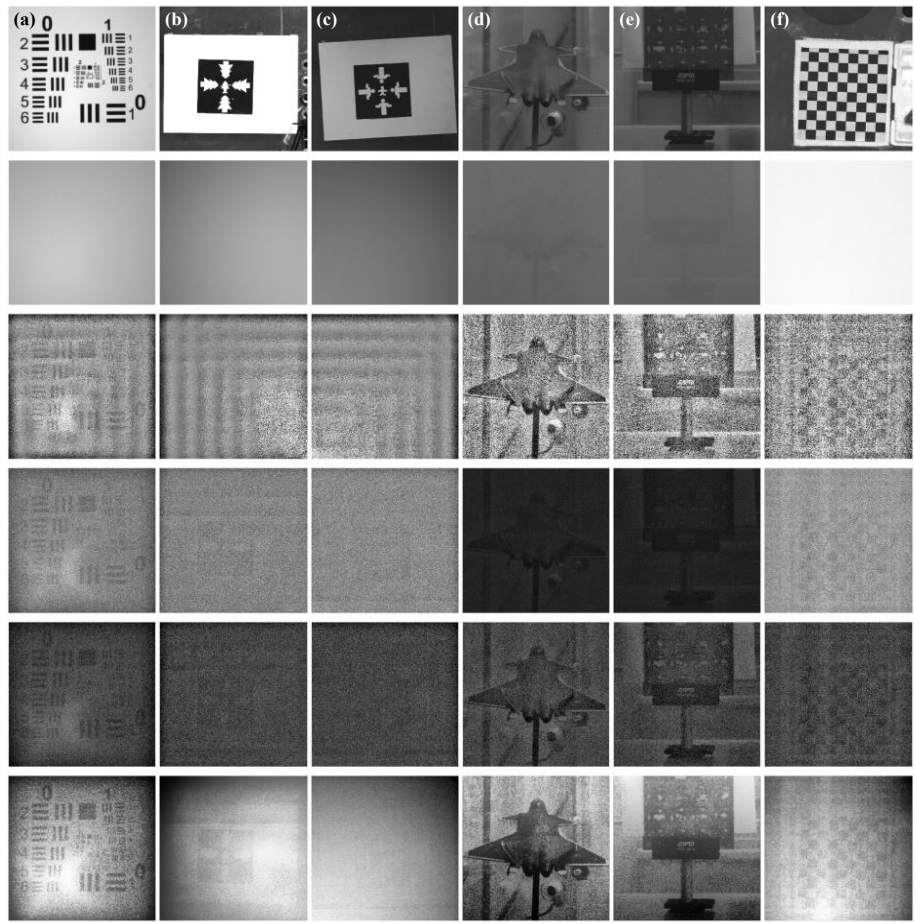

**Fig. 5.** Comparison of different image enhancement algorithms in extremely low visibility condition inside the fog chamber. From left to right are clear image, original image, CLAHE, SSR, AMSR, and our algorithm respectively. (a)-(f) demonstrates different targets.

For quantitative analysis, the majority of index of our algorithm is higher. However, in Fig. 5 (d), (e) SSR exhibits a better score, this result is of an inadequate exposure of the clear images in both scenarios, which coincidently aligns the dark result coming from SSR. In normal cases with appropriate exposure-time for clear images, HMHE always exhibits higher scores.

**Table 1.** Qualitative comparisons between IE, SSIM, PSNR, and CORR of different algorithms in indoor experiment.

| Figure | IQA | CLAHE | SSR | AMSR | Ours |
|--------|-----|-------|-----|------|------|
| 5(a) | IE | 7.45209 | 6.85442 | 6.73334 | **7.75753** |
|  | SSIM | 0.08023 | 0.10667 | 0.08196 | **0.19229** |
|  | FSIM | 0.42930 | 0.47389 | 0.47204 | **0.48000** |
|  | CORR | 0.11995 | 0.14259 | 0.13011 | **0.17085** |
| 5(b) | IE | 7.38049 | 6.74672 | 6.70964 | **7.73492** |
|  | SSIM | 0.08243 | 0.07527 | 0.06774 | **0.31135** |
|  | FSIM | 0.60550 | 0.59779 | 0.59932 | **0.61148** |
|  | CORR | 0.23337 | 0.12780 | 0.13975 | **0.26364** |
| 5(c) | IE | 7.33908 | 6.75464 | 6.70898 | **7.61570** |
|  | SSIM | 0.07678 | 0.07979 | 0.08278 | **0.29756** |
|  | FSIM | 0.59185 | 0.60162 | 0.61475 | **0.62179** |
|  | CORR | 0.16030 | 0.09169 | 0.10638 | **0.27352** |
| 5(d) | IE | **7.95973** | 4.86909 | 6.70503 | 7.68406 |

| Figure | IQA | CLAHE | SSR | AMSR | Ours |
|---|---|---|---|---|---|
| | SSIM | 0.02726 | **0.41822** | 0.12946 | 0.23182 |
| | FSIM | 0.39348 | **0.80907** | 0.67776 | 0.57946 |
| | CORR | 0.31503 | 0.35087 | 0.36047 | **0.72899** |
| 5(e) | IE | **7.95370** | 4.98626 | 6.73997 | 7.60437 |
| | SSIM | 0.02071 | **0.43820** | 0.11987 | 0.33159 |
| | FSIM | 0.41044 | **0.78605** | 0.66660 | 0.68101 |
| | CORR | 0.28866 | 0.40698 | 0.42350 | **0.66119** |
| 5(f) | IE | **7.91214** | 6.54692 | 6.69851 | 7.56860 |
| | SSIM | 0.03088 | 0.11613 | 0.09479 | **0.21201** |
| | FSIM | 0.37762 | 0.48803 | 0.47302 | **0.50114** |
| | CORR | 0.09291 | 0.10340 | **0.10464** | 0.05351 |

The comparative analysis of experimental data from outdoor settings is depicted in Fig. 6. This figure illustrates six distinct outdoor scenarios, each represented in columns (a) to (f). Sequentially from top to bottom, each set presents the original images followed by the results processed using CLAHE, SSR, AMSR, and the algorithm proposed herein. Analogous to indoor findings, the restoration quality of CLAHE sees improvement due to the diminished irregularities in lighting. Despite SSR's capacity to restore images, it continues to yield comparatively low contrast. AMSR here demonstrates even lower contrast than SSR. HMHE successfully recovers critical visual elements in outdoor scenes. Quantitative assessments are shown in Table 2, where the best scores are highlighted with bold text. It can be seen in outdoor conditions, HMHE is of advantageous qualitative score.

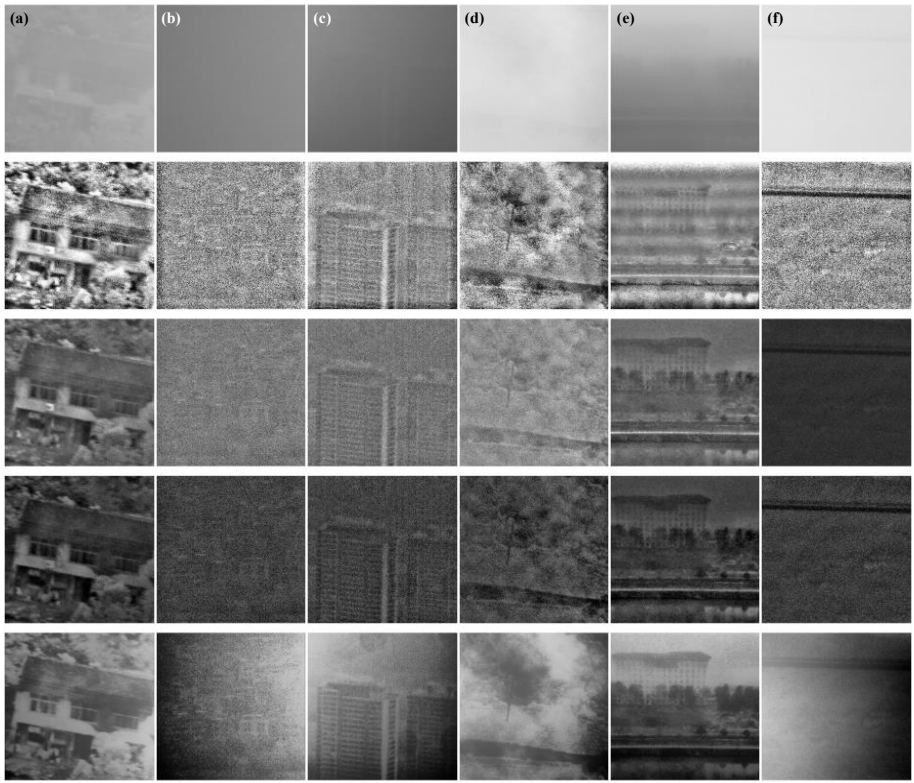

**Fig. 6.** Comparisons of different image enhancement algorithms in the outdoor experiment. (a)-(f) are in different scenes. From up to below are respectively the original images, CLAHE, SSR, AMSR, and ours.

**Table 2.** Qualitative comparisons between IE, SSIM, PSNR, and CORR of different algorithms in outdoor experiment.

| Figure | IQA | CLAHE | SSR | AMSR | Ours |
|---|---|---|---|---|---|
| 6(a) | IE | **7.92424** | 6.76166 | 6.64734 | 7.50889 |
| | SSIM | 0.13068 | 0.41596 | 0.30377 | **0.62899** |
| | FSIM | 0.33459 | 0.65163 | 0.66496 | **0.68451** |
| | CORR | 0.50456 | 0.53852 | 0.54223 | **0.94058** |
| 6(b) | IE | 7.87551 | 6.79691 | 6.73705 | **7.88411** |
| | SSIM | 0.06829 | 0.19673 | 0.15419 | **0.25255** |
| | FSIM | 0.26334 | 0.49349 | 0.47663 | **0.51187** |
| | CORR | 0.30587 | 0.30300 | 0.31409 | **0.94419** |
| 6(c) | IE | 7.62235 | 6.78204 | 6.72882 | **7.80304** |
| | SSIM | 0.10018 | 0.19934 | 0.16949 | **0.38536** |
| | FSIM | 0.47190 | 0.64911 | 0.64058 | **0.68164** |
| | CORR | 0.27570 | 0.17789 | 0.18539 | **0.95386** |
| 6(d) | IE | **7.82762** | 6.58170 | 6.72037 | 7.57464 |
| | SSIM | 0.09130 | 0.24309 | 0.12753 | **0.53592** |
| | FSIM | 0.41575 | 0.67563 | 0.62096 | **0.72060** |
| | CORR | 0.45556 | 0.35883 | 0.36952 | **0.92809** |
| 6(e) | IE | 7.07433 | 6.41182 | 6.49151 | **7.70322** |
| | SSIM | 0.36827 | 0.41095 | 0.28306 | **0.54742** |
| | FSIM | 0.57004 | 0.66446 | 0.63293 | **0.68455** |
| | CORR | 0.24727 | 0.22330 | 0.24671 | **0.95100** |
| 6(f) | IE | **7.93681** | 5.46808 | 6.58479 | 7.86451 |
| | SSIM | 0.05784 | 0.24139 | 0.11950 | **0.52467** |
| | FSIM | 0.23767 | 0.76811 | 0.50903 | **0.79610** |
| | CORR | 0.57649 | 0.62285 | 0.61980 | **0.82934** |

## 4. Conclusions

This study introduces a novel algorithm designed to substantially enhance image contrast under conditions of extremely low visibility. The algorithm strategically eliminates redundant low-frequency information through SDIF before proceeding to image enhancement. The filter parameter is adjudicated based on the structural similarity between the post-filtered and original images, with images deemed devoid of valuable information upon a significant shift in similarity rates. Post-filtering, images may exhibit a grayscale banding effect, which is effectively mitigated through deliberate disturbance introduction, thus ensuring an improved visual presentation. The algorithm concludes with a comprehensive global enhancement of the image, amplification of the interested high-frequency components, and recombination of low-frequency components. Through rigorous indoor and outdoor experimental validation, the algorithm demonstrates a pronounced capability to enhance image contrast significantly, thereby maximizing the visibility of objects under severely limited visibility conditions. Our result shows potential uses for further application in low visibility imaging.

**Author Contributions:** Conceptualization, L.C.; methodology, L.C.; software, L.C. and J.L.; validation, L.C. and J.L.; formal analysis, L.C.; investigation, L.C. and Q.B.; resources, Y.L. and J.Z.; data curation, L.C. and J.L.; writing—original draft preparation, L.C.; writing—review and editing, L.C., J.L., Q.B., Y.L., and J.Z.; visualization, L.C.; supervision, Y.L.; project administration, Y.L.; funding acquisition, Y.L. and J.Z. All authors have read and agreed to the published version of the manuscript.

**Funding:** This work was supported by the Guangdong Major Project of Basic and Applied Basic Research (2020B0301030009); National Natural Science Foundation of China (61991452); National Key Research and Development Program of China (2022YFA140430402).

**Institutional Review Board Statement:** Not applicable.

**Informed Consent Statement:** Not applicable.

**Data Availability Statement:** The original data presented in the study are openly available in Chen-Lubbon/Image-enhencement-code (github.com).

**Conflicts of Interest:** The authors declare no conflicts of interest.